\documentclass[conference]{IEEEtran}
\IEEEoverridecommandlockouts

\usepackage{amsmath,amssymb,amsfonts}
\usepackage{comment}
\usepackage{algorithmic}
\usepackage{algorithm}
\usepackage{flushend}
\usepackage{balance}
\usepackage{cite}
\usepackage{textcomp}
\usepackage[linesnumbered,lined,boxed,commentsnumbered,algo2e]{algorithm2e}
\usepackage{tabu}

\usepackage[dvipsnames]{xcolor}
\newcommand{\report}[1]{{}}
\newcommand{\paper}[1]{{#1}}

\newcommand{\hamid}[1]{\textcolor{RubineRed}{#1}}
\newcommand{\hamidDone}[1]{#1}

\newcommand{\gregDone}[1]{#1}
\newcommand{\Mahshid}[1]{#1}

\usepackage{hyperref}
\usepackage{textcomp}
\usepackage{xcolor}
\hypersetup{
     colorlinks = true,
     linkcolor = blue,
     anchorcolor = blue,
     citecolor = green,
     urlcolor = blue
     }

\def\BibTeX{{\rm B\kern-.05em{\sc i\kern-.025em b}\kern-.08em
    T\kern-.1667em\lower.7ex\hbox{E}\kern-.125emX}}
\ifCLASSINFOpdf
  \usepackage[pdftex]{graphicx}
\else
\fi

\ifCLASSOPTIONcompsoc
    \usepackage[caption=false, font=normalsize, labelfont=sf, textfont=sf]{subfig}
\else
\usepackage[caption=false, font=footnotesize]{subfig}
\fi

\emergencystretch=\maxdimen
\begin{document}
%
\title{Efficient and Effective Generation of Test Cases for Pedestrian Detection -- Search-based Software Testing of Baidu Apollo in SVL}

\author{\IEEEauthorblockN{Hamid Ebadi\IEEEauthorrefmark{1},
Mahshid Helali Moghadam\IEEEauthorrefmark{2},
Markus Borg\IEEEauthorrefmark{2}\IEEEauthorrefmark{4}, 
Gregory Gay\IEEEauthorrefmark{3},
Afonso Fontes\IEEEauthorrefmark{3} and
Kasper Socha\IEEEauthorrefmark{4}}
\IEEEauthorblockA{\IEEEauthorrefmark{1}Infotiv AB, Sweden}
\IEEEauthorblockA{\IEEEauthorrefmark{2}RISE Research Institutes of Sweden, Sweden}
\IEEEauthorblockA{\IEEEauthorrefmark{3}Chalmers and the University of Gothenburg, Sweden}
\IEEEauthorblockA{\IEEEauthorrefmark{4}Lund University, Sweden}}

\maketitle

\begin{abstract}
With the growing capabilities of autonomous vehicles, there is a higher demand for sophisticated and pragmatic quality assurance approaches for machine learning-enabled systems in the automotive AI context. The use of simulation-based prototyping platforms provides the possibility for early-stage testing, enabling inexpensive testing and the ability to capture critical corner-case test scenarios. Simulation-based testing properly complements conventional on-road testing. However, due to the large space of test input parameters in these systems, the efficient generation of effective test scenarios leading to the unveiling of failures is a challenge. 

This paper presents a study on testing pedestrian detection and emergency braking system of the Baidu Apollo autonomous driving platform within the SVL simulator. We propose an evolutionary automated test generation technique that generates failure-revealing scenarios for Apollo in the SVL environment. Our approach models the input space using a generic and flexible data structure and benefits a multi-criteria safety-based heuristic for the objective function targeted for optimization. This paper presents the results of our proposed test generation technique in the 2021 IEEE Autonomous Driving AI Test Challenge. In order to demonstrate the efficiency and effectiveness of our approach, we also report the results from a baseline random generation technique. Our evaluation shows that the proposed evolutionary test case generator is more effective at generating failure-revealing test cases and provides higher diversity between the generated failures than the random baseline.
\end{abstract}


\section{Introduction}
The capabilities of autonomous vehicles have increased remarkably in recent years. A self-driving car is arguably the most tangible example of what the European Commission (EC) defines as an Artificial Intelligence (AI) system~\cite{ec_ai_def}. From an AI perspective, the automotive industry has successfully harnessed the disruptive potential of machine learning over the last decade. Driven by the availability of big data and computing power, deep neural networks (DNNs) have enabled new levels of vehicular perception. However, performing effective quality assurance of systems that rely on DNNs requires a paradigm shift~\cite{borg2019safely}. No longer do human engineers explicitly express all logic of the system in source code. Instead, DNNs are trained using enormous quantities of manually annotated data and perform actions probabilistically based on patterns observed in that data. The research community has put substantial effort into making DNN-based systems trustworthy in the automotive AI context, spurring major R\&D projects and global safety standardization efforts.

The concept of Trustworthy AI receives particular attention in the EC's AI Strategy~\cite{ec_ai_strategy}. EC defines AI systems as ``software (and possibly also hardware) systems designed by humans that, given a complex goal, act in the physical or digital dimension by perceiving their environment through data acquisition, interpreting the collected structured or unstructured data, reasoning on the knowledge, or processing the information, derived from this data and deciding the best action(s) to take to achieve the given goal''~\cite{ec_ai_def}. Novel ways to test AI systems, including autonomous vehicles, are urgently needed---and the research community has taken up the challenge~\cite{zhang2020machine,riccio2020testing}.

The use of virtual prototyping platforms for automotive software engineering has rapidly grown in recent years~\cite{bock2019status}. The use of virtual methods allows testing and validation at early development stages, which leads to fewer development cycles and faster time-to-market. Simulation-based testing is required to complement conventional on-road testing due to severe drawbacks in the use of on-road testing~\cite{koopman2016challenges}, i.e., system testing on public roads is costly and does not scale to the quantity of scenarios needed---in addition, it can be dangerous to provoke a critical situation on the road. Testing autonomous vehicles in simulators is fundamental to quality assurance in the automotive sector---as indicated in the evolving standard ISO 21448 Safety of the Intended Functionality~\cite{international_organization_for_standardization_road_2019}.

\report{
\subsection{Test Organization and Roles}
\begin{itemize}
\item Hamid Ebadi -- Team leader and lead designer/developer
\item Mahshid Helali Moghadam -- Test designer/developer
\item Markus Borg --  Test designer
\item Gregory Gay -- Test designer
\item Afonso Fontes -- Test designer
\item Kasper Socha -- Reviewer
\end{itemize}
}

Efficient and effective testing in simulated environments requires sophisticated approaches to automatically generating test cases. Several authors have demonstrated that search-based software test generation (SBST)~\cite{mcminn2011search} is a feasible approach to generate critical test scenarios in the automotive context~\cite{ben_abdessalem_testing_2016,ben_abdessalem_testing_2018,gambi_generating_2019,borg2021digital,Moghadam2021Deeper}, i.e., test scenarios that lead to the violation of safety requirements. SBST formulates test input selection as a search problem, where optimization algorithms attempt to systematically identify the test input that meet goals of interest. Given a scoring function denoting \textit{closeness to the attainment of those goals}---called \textit{objective function}---optimization algorithms can sample from a large and complex set of test inputs as guided by a chosen sampling strategy (a \textit{metaheuristic}---in our case, a genetic algorithm)~\cite{mcminn2011search}. 

In the 2021 IEEE Autonomous Driving AI Test Challenge competition, our contribution, \texttt{ScenarioGenerator}, uses SBST to generate test scenarios that cause the Baidu Apollo's autonomous driving platform to fail. \hamidDone{While  different scenarios can be tested using \texttt{ScenarioGenerator}, for the purpose of this work, we assume a scenario with a pedestrian crossing a street} with the following high-level safety goal: ``The ego car shall not crash into pedestrians on collision course.'' We refer to any crashes between an ego car and pedestrians as a safety violation \Mahshid{or failure}.

Our work relies on a test strategy involving the following steps of simulation-based automotive testing using SBST. We:
\begin{enumerate}
\item Build a scene in the virtual environment.
\Mahshid{\item Define the parameters involved in creating a varied set of test cases.}
\Mahshid{\item Define ranges for each parameter, representing the test input space to explore.}
\item Define an objective function that measures the \textit{quality} of a generated test case, in terms of its potential to demonstrate a safety violation. In our case, lower scores indicate more dangerous scenarios.
\Mahshid{\item Apply a genetic algorithm to generate test cases that minimize the objective function, leading to safety-critical scenarios.} 
\end{enumerate}

To accomplish this, we first import a pre-existing map into the SVL Visual Scenario \Mahshid{Editor} and create \Mahshid{an initial} movement path for a pedestrian using fixed waypoints---a set of coordinates (points) showing the initial path of the pedestrian's movements. \Mahshid{Then, during the simulation, in the designed scene,} the ego car moves forward toward a target and a pedestrian crosses the road from the right.

The proposed evolutionary test case generation formulates the search space using a generic \textit{noise vector} data structure and minimizes a multi-criteria objective function that combines (1) distances between the ego car and other road agents, (2) the distance of the journey taken by the ego car towards the target, and (3), the number of accidents detected.\Mahshid{ Using the noise vector, as a generic and flexible structure for representing the search space of the problem, facilitates the use of a wide variety of search algorithms.  
This \report{report}\paper{paper} presents the results of our proposed test case generation technique in the 2021 IEEE Autonomous Driving AI Test Challenge. To  provide the comparative results and demonstrate the efficiency and effectiveness of our evolutionary text case solution, we also compare our results to random generation of test scenarios.}

\report{
The rest of the report is organized as follows: Section \ref{sec:TCGenerationSBST} presents the details of the proposed search-based test case generation solution, and Section~\ref{sec:generate} elaborates deeply on the empirical evaluation including the research method, test scenario execution, the experimental results, and the discovered problems and issues during the experimentation. 
Section~\ref{sec:summary} summarizes our findings in the light of the importance of simulation-based testing of autonomous vehicles and potential research directions for the future work.
}

\paper{
The rest of the paper is organized as follows: Section \ref{sec:TCGenerationSBST} presents the details of the proposed search-based test case generation approach. Section~\ref{sec:generate} elaborates on the empirical evaluation, including the research method, test scenario execution and experiment setup, results, and threats to the validity of the results. 
Section \ref{sec:relatedwork} presents an overview of related work, and Section~\ref{sec:summary} summarizes our findings in light of the importance of simulation-based testing of autonomous vehicles and potential research directions for future work.
}

\section{Search-based Test Case Generation} \label{sec:TCGenerationSBST}

This section presents how we use an evolutionary search-based technique to generate test cases. Since each scenario takes a few seconds to execute, it is not feasible to try all possible test scenarios. \Mahshid{Our approach is to adopt a generic data structure, i.e., a data vector called a ``noise vector'', to represent the test input domain for producing test scenarios. Each element of this vector represents a parameter that defines a test scenario, e.g., waypoints, illumination, and weather. The values of these parameters do not lie within the same range, so to bind the values within a specific range, the input representation also scales the concrete real values to values within the range $[-1, +1]$. The values in the noise vector are manipulated by the search algorithm to produce test cases. In our approach, we use a genetic algorithm to explore the search space and produce test cases that are judged as more valuable using an objective function based on potential pedestrian collisions.} 

\subsection{Scenario Creation and Manipulation} \label{sec:manipulate}

\Mahshid{We use SVL Visual Scenario Editor as the first step to create a basic scheme of the test scenarios that are going to be executed by SVL simulator.} SVL Visual Scenario Editor is a GUI application that can be used to create basic scenarios specifying where agents (pedestrians, vehicles, ego vehicle, etc.) are positioned in a map and the basic scheme of the path that they should take through the map, which is specified in the form of waypoints. 

The basic scenario is created and exported from SVL Visual Scenario Editor to SVL simulator. This scenario is then manipulated by \texttt{ScenarioGenerator} to produce new test scenarios. 
In \texttt{ScenarioGenerator}, a derived test scenario is specified by a vector of real numbers, the \textit{noise vector}, with values between $-1$ and $+1$. 

\subsection{Scenario specification} 
\Mahshid{A test scenario is defined as a set of parameters used for test scenario generation, i.e., modeling the test inputs, which is shown as follows:}
\begin{equation}
\begin{split}
TS= \langle S_1, S_2, \cdots, S_m \rangle  \ , \  &{R_i}_{min} \leq S_i \leq {R_i}_{max}\\
&{R_i}_{min}, {R_i}_{max} \in \mathbb{R}
\end{split}
\end{equation}
\Mahshid{Where $TS$ indicates a test scenario and $S_i$ denotes a test input parameter. The values of the test input parameters often vary over different ranges. ${R_i}_{min}$ and ${R_i}_{max}$ represent the upper and lower boundaries of the value range for parameter $S_i$.}  

\gregDone{For example, the scenario may define a variable $S_{tod}$ representing the time of day. In the base scenario, the time of day may be defined as 12:00. ${R_{tod}}_{min}$ and ${R_{tod}}_{max}$ are used to limit the \textit{change} in this value in a generated test scenario (e.g., values of $-5$ and $5$ would allow the time to vary from 7:00 to 17:00).} The values of parameters representing the positions of the agents would have different ranges---e.g., the position points in a path that the vehicle takes may change by $\pm$2 (meters).

\subsection{Noise vector} 
\Mahshid{The proposed representation for a test case is a vector, which is defined as follows:}
\begin{equation}
noise\_vector = \langle N_1, N_2, \cdots, N_m \rangle\ , \  -1 \leq N_i \leq +1 
\end{equation}
\Mahshid{where each element, $N_i$, corresponds to a test input parameter, $S_i$, and the values of components of the noise vector are scaled to values in $R$ using a linear scaling function to create a test scenario, $TS$.}
\begin{equation}
S_i = (N_i+1 ) \times ({R_i}_{max} - {R_i}_{min}) / 2  + {R_i}_{min}
\end{equation}
This transformation allows the use of a generic representation that can be uniformly manipulated by the test generator without detailed knowledge of each input parameter. All elements of the noise vector fall within the range $[-1, +1]$, and are scaled appropriately using ${R_i}_{min}$ and ${R_i}_{max}$ for that $S_i$.

\gregDone{For example, a noise vector value of 0.5 for the entry representing the time of day, $S_{tod}$, would result in the following concrete value in a test case: $S_{tod} = (0.5 + 1) \times (17 - 5)/2 + 5 = 1.5 \times 6 + 5 = 14$, or 14:00.}

\subsection{Objective Function} \label{sec:objfunc}

In order to generate valuable test scenarios, we must identify scenarios that are more likely to lead to safety violations. Safety violations can occur then the ego car moves toward its target at a reasonable speed. Specifically, the objectives to be optimized are as follows:

\begin{itemize}
	\item The total distance\footnote{Euclidean distance
\begin{align*}
d(p1,p2) = \sqrt {\left( {p1_x - p2_x } \right)^2 + \left( {p1_y - p2_y }\right)^2  + \left( {p1_z - p2_z }\right)^2  })
\end{align*}}
        of the ego vehicle from other non-ego traffic during scenario execution. This objective should be \textit{minimized}---we want to examine ego vehicle behavior in potentially dangerous scenarios.
        \begin{equation}
        \begin{split}
        ego\_agents\_distance =&\\
        \sum_{agent \in agents}\sum_{s \in (1, ..., steps)}d(ego.pos_s , agent.pos_s)&
        \end{split}
        \end{equation}
	\item The total distance of the journey. This should be \textit{maximized}, as longer journeys are preferred.
	\begin{equation}
	journey\_distance = d(ego.pos_{1}, ego.pos_{finalstep})
    \end{equation}
        \item $acc$ : the number of accidents. This should also be \textit{maximized}, as we seek failures in ego vehicle behavior. 
\end{itemize}

\Mahshid{Since the aforementioned objectives do not conflict with each other, we merge them to form a single objective function. This function is \textit{minimized}---lower scores are preferred. The objective function that we seek to minimize is defined as:}
\begin{equation}
E = ego\_agents\_distance - journey\_distance - 1000 \times acc
\end{equation}
\Mahshid{We put high values on the number of accidents, as we are interested in generating test scenarios leading to crashes.}


\subsection{Search Algorithm}

\hamidDone{It is not possible to execute every possible test scenario that can be defined by an instance of the noise vector.} Instead, we seek a systematic means to sample from the space of possible scenarios in \textit{search} of those that could lead to safety violations. This can be done by using an optimization algorithm to sample the space, as guided by the objective function. 



The optimization algorithm used \hamidDone{to minimize the objective function} is a Genetic Algorithm (GA). 
Genetic Algorithms are modeled on the evolution of a population over time. Initially, a random population of solutions (\textit{noise\_vector} instances) is generated. Then, at each generation, a new population is formed based on the best solutions resulting from the previous generations of evolution. This population is formed by:
\begin{itemize}
    \item Identifying good solutions using \textit{tournament selection}, where a subset of the population is selected at random and the best member of the subset is identified.
    \item Breeding ``child'' solutions by combining elements of ``parent'' solutions through \textit{crossover}, where the child solutions are formed by selecting genes (elements) from each parent solution.
    \item Introducing \textit{mutations} into the population by making small, random adjustments to solutions.
\end{itemize}
Tournament selection is performed to identify parent solutions, then crossover and mutation are performed at user-set probabilities. Either, or both, may be applied to transform the identified solutions. Finally, the resulting solutions are added to the new population. This process continues until a new population is formed. The objective function is calculated for each member of this population, and the score is stored for that solution. This process is performed each generation, until a user-set number of generations has been exhausted. At the end, the best solutions are returned.  

In our case, we have three objectives---$ego\_agents\_distance$, $journey\_distance$, and $acc$, \Mahshid{which have been merged into a single formula. 
Tournament selection picks the best solution among the solutions in each tournament. The number of individuals participating in each tournament denotes the size of the tournament.}    
\Mahshid{In our approach, we omit the crossover operation, as the noise vector contains the values for the parameters of the test scenarios in a certain order, and crossover could violate this ordering. Instead, we apply mutation with a high probability. We use \textit{Polynomial Bounded mutation}, as proposed and implemented in NSGA-II \cite{deb2002fast}. It is a bounded mutation operation for real-valued parameters and uses a polynomial function for the probability distribution. It uses a parameter, $eta$ indicating the \textit{crowding degree} of the mutation, which is used to encourage diversity in the resulting population. A high $eta$ yields a mutant resembling the original solution, while a small value for $eta$ produces a solution more divergent from the original.
The GA algorithm used for generating test scenarios is configured as presented in Algorithm \ref{Algorithm:GA}.}

\begin{algorithm}[h]
\SetAlgoLined
\begin{flushleft}
\caption{GA for Test Scenario Generation}\label{Algorithm:GA}
\textrm{Initialize population with solutions from random seeds};\\
\textrm{Evaluate the population};\\
\Repeat{meeting the stopping criteria (reaching the maximum number of generations or other limitations specified in the test budget)}{
        \textrm{1. Select offspring using Tournament Selection with replacement};\\ 
        \textrm{2. Mutate the resulting offspring using \textit{Polynomial Bounded mutation} operation with a certain probability (mutation rate = 0.95)};\\
        \textrm{3. Evaluate the offspring using the objective function.}}
\end{flushleft}
\end{algorithm}

\begin{figure*}[!t]
  \centering
  \includegraphics[width=.85\textwidth]{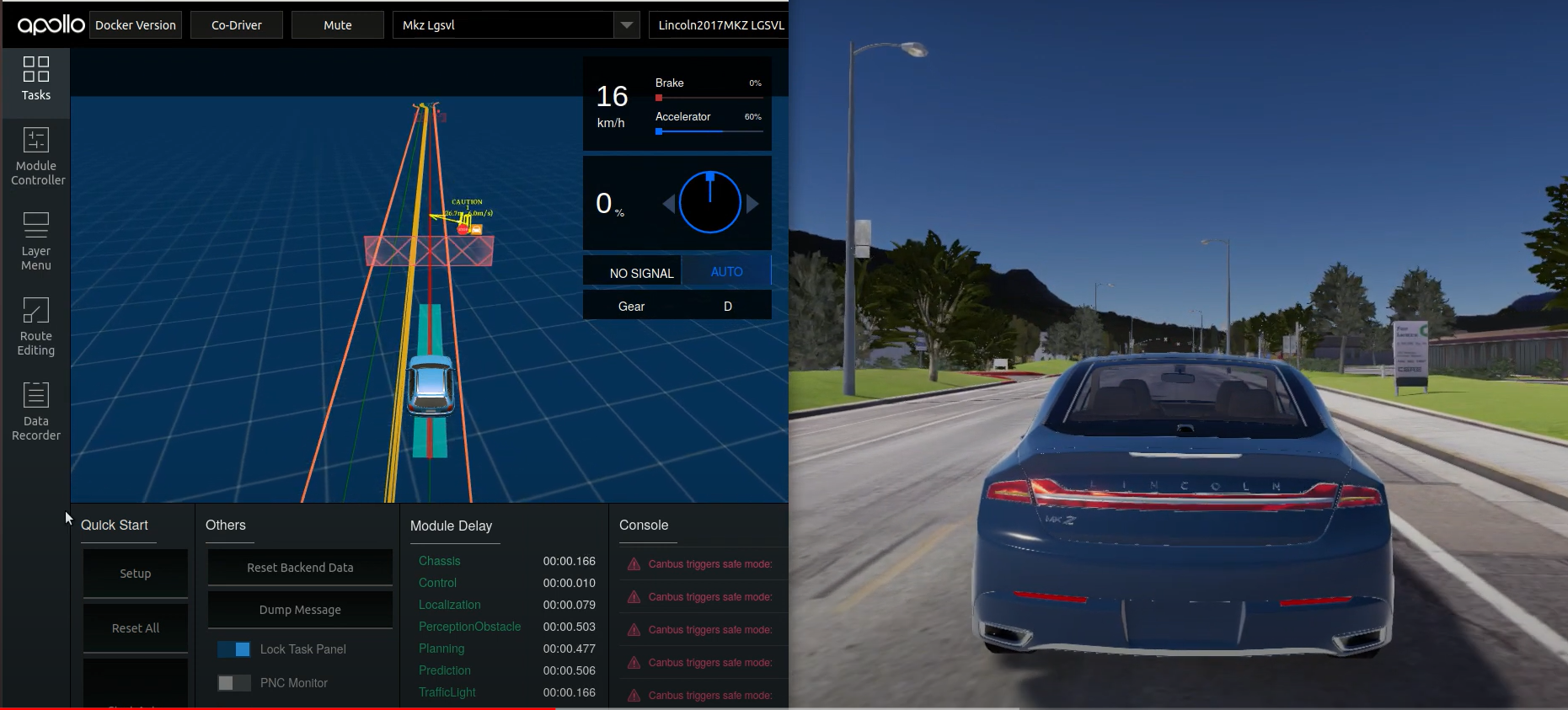}
  \caption{Overview of the experimental setup.}
  \label{fig:expSetup}
\end{figure*}

\section{Implementation and Empirical Evaluation}\label{sec:generate}

\Mahshid{We perform an empirical evaluation of the proposed test case generation technique, \texttt{ScenarioGenerator}\footnote{Available from \url{https://github.com/ebadi/ScenarioGenerator}.} by running experiments on our experimental setup} on a desktop PC with the following specifications:
\begin{itemize}
\item Ubuntu version 18.04
\item Intel Core i7-10700K CPU @ 3.80GHz × 16
\item 32GB RAM
\item GeForce RTX 2070 SUPER/PCIe/SSE2
\item SVL simulator 2021.1 (linux64) with modular testing setup (3D Ground Truth sensor and Signal sensor publish ground truth perception data to Apollo via CyberRT bridge)
\item Baidu Apollo (r6.0.0 branch)
\end{itemize} 
\Mahshid{The experiments are simulations that are controlled by a Python scenario runner which uses our test case generation technique for generating the scenarios in the simulation environment. Baidu Apollo is the autonomous driving software platform that controls the ego vehicle. It connects to the simulator through its customized bridge and drives the ego vehicle (Fig. \ref{fig:expSetup}).} 

\Mahshid{We design a set of experiments to assess the efficiency and effectiveness of the proposed test case generation for testing Apollo in the SVL simulation environment. Pedestrian detection and proper responding is the target use case of Apollo in our experiments. For a comparative analysis, we also report results from a random testing technique as a baseline approach. In random testing, the test cases are generated randomly, which means that the set of noise vector instances are generated by setting the test input parameters to random values within the allowed range. The target is to generate the highest number of diverse valid test cases leading to failures, i.e., collisions between the ego vehicle and pedestrians. We use the following quality criteria for evaluating the proposed test case generation technique:
\begin{itemize}
    \item \textbf{Detected Failures:} The number of test cases that lead to a collision.
    \item \textbf{Failure Diversity:} The dissimilarity between the generated test cases leading to failures. We are interested in generating diverse test cases, as triggering similar failures leads to waste of the test budget, e.g., computation resources. To measure failure diversity, we use the \textit{Euclidean distance} between failure-revealing noise vectors.
\end{itemize}
}

\subsection{Test Scenario Execution}\label{sec:TestExecution}

The testing budget (including, e.g., execution time) is a limited resource. While not as expensive to perform as on-road testing, running test scenarios in simulators also takes time. In our experiments, each scenario takes about 10 seconds to execute and evaluate. Therefore, for the purpose of this competition, we set the limit for the number of simulation executions to 200 in the Genetic Algorithm. This would correspond, for example, to 20 generations with a population size of 10. 
\report{
Later, in Section \ref{Sec:problems} we will discuss emerging issues related to SVL simulator that limit the possibility of running long experiments.}

\paper{
}
 \report{ You can see a video of the  \texttt{ScenarioGenerator} in action in
 \hamid{\href{https://www.youtube.com/watch?v=2LEu7W-j_Eg}{this link}}.

In the developed test case generator tool, the user controllable parameters for test scenario creation and manipulation are as follows:
\begin{itemize}
	\item \texttt{{-}{-}input}: Path to the JSON file created by SVL Visual Scenario Editor.
	\item \texttt{{-}{-}action}: Specifying what strategies have to be used for scenario generation. Currently \texttt{differential\_evolution}, \texttt{powell}, \texttt{genetic\_algorithm} and \texttt{random} strategies are supported.  You can also replay one specific scenario by passing the json file and setting the action to \texttt{replay}.
	\item \texttt{{-}{-}vector}: This can optionally be used in combination with \texttt{replay} action to add an specific noise vector to a scenario.   In this mode, in addition to all the previous parameters, one specific vector of noise is given to be played.

	\item \texttt{{-}{-}des-forward-right}: destination relative to where the ego vehicle is placed. 
	\item \texttt{{-}{-}seed}: Randomness seed
	\item \texttt{{-}{-}steps}: The duration of the scenario in second. 
	\item \texttt{{-}{-}pos-noise-range-xz}: Acceptable range of change in the values for the position of each waypoint (x,z).
	\item \texttt{{-}{-}color-noise-range-rgb}: Acceptable range of change in the values for agent color (r, g, b).
	\item \texttt{{-}{-}weather-noise-range}: Acceptable range of change in the values for the weather in the simulation (rain, fog, wetness, cloudiness, road damages) .
	\item \texttt{{-}{-}time-max-noise}: Acceptable range of change in the values for the day time in the simulator.
	\item \texttt{{-}{-}speed-max-noise}: Acceptable range of change in the values for agent speed.
\end{itemize}}



\paper{

\begin{figure*}[h]
\centering
\subfloat[Number of detected failures.]{
    \framebox{\includegraphics[width=0.32\linewidth]{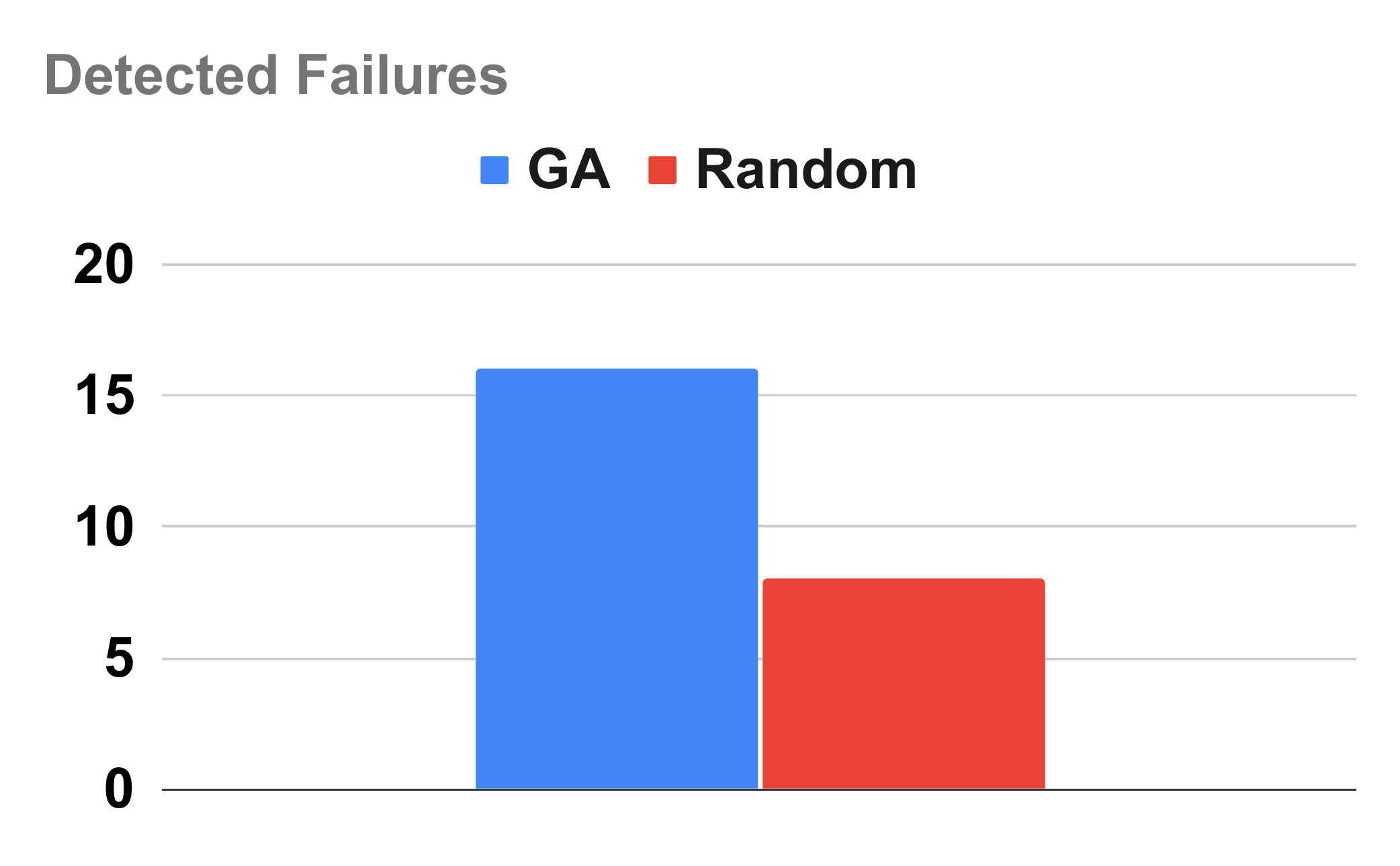}} } 
\subfloat[Objective values for the average journey distance during failure-revealing test cases.]{
    \framebox{\includegraphics[width=0.32\linewidth]{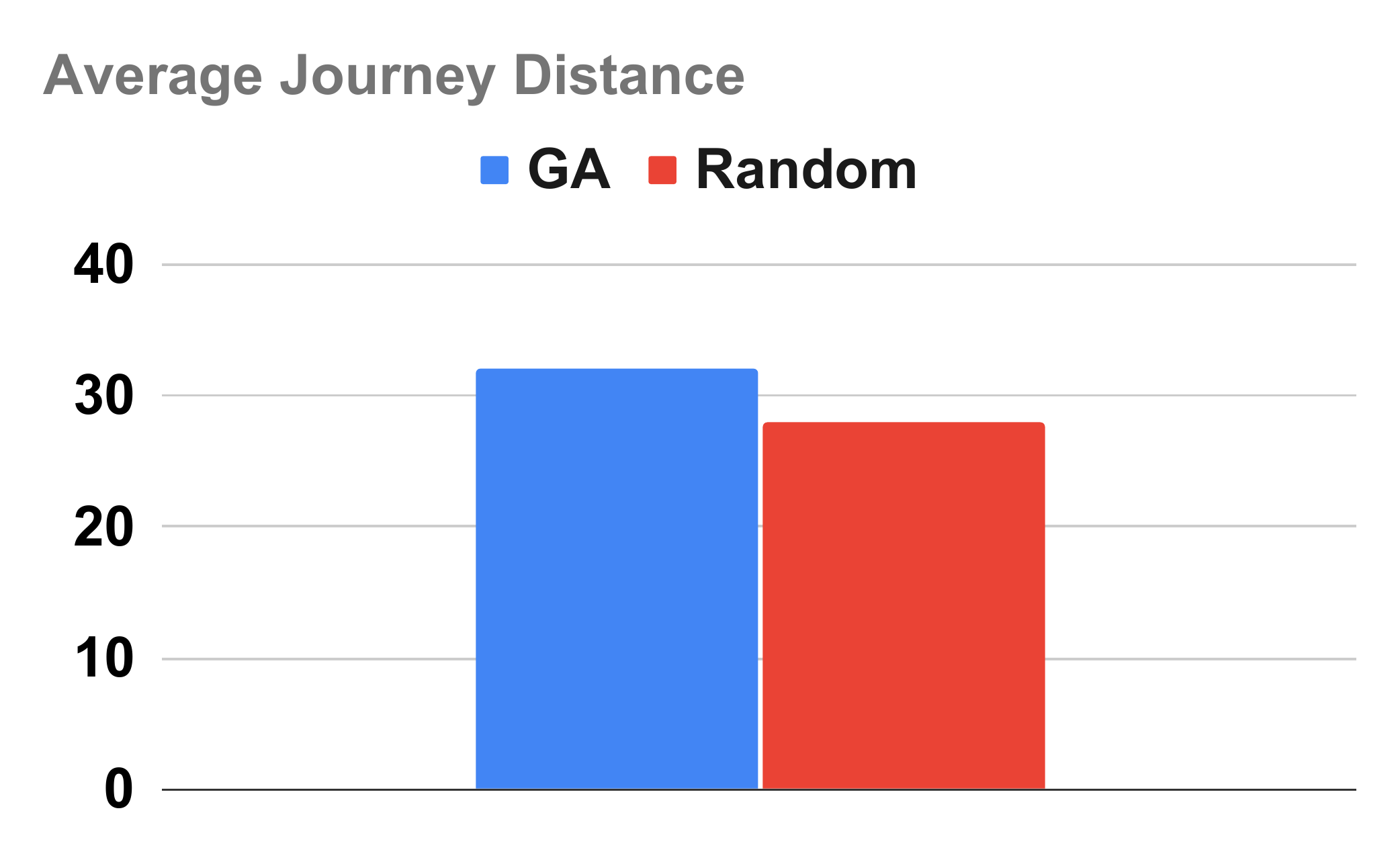}}}
\subfloat[Objective values for average distance from ego car during failure-revealing test cases.]{
    \framebox{\includegraphics[width=0.32\linewidth]{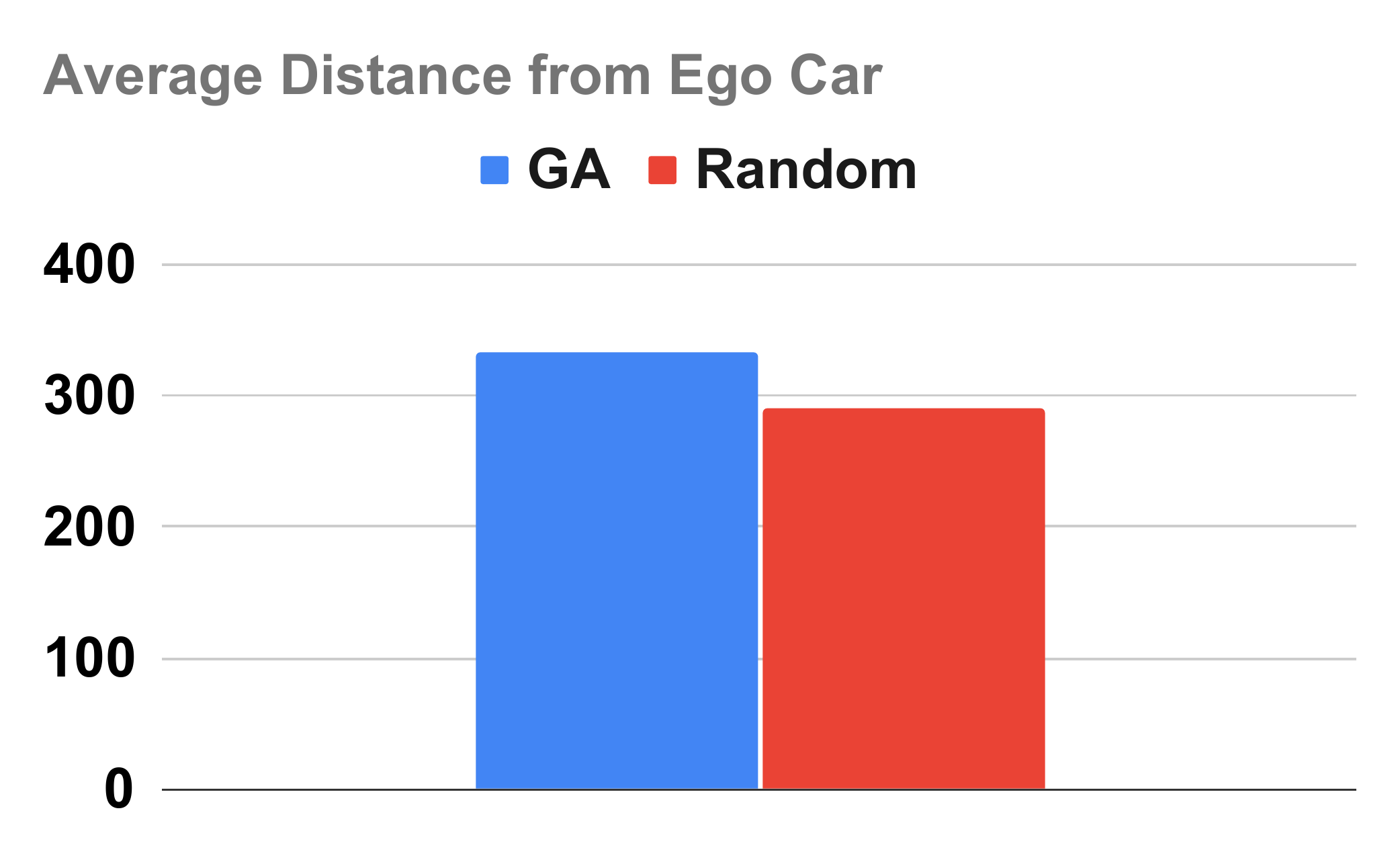}} }
 \caption{Comparisons between GA and random generation.} 
  \label{fig:failure_data}
\end{figure*}

\begin{figure}[!t]
  \centering
  \includegraphics[width=.95\columnwidth]{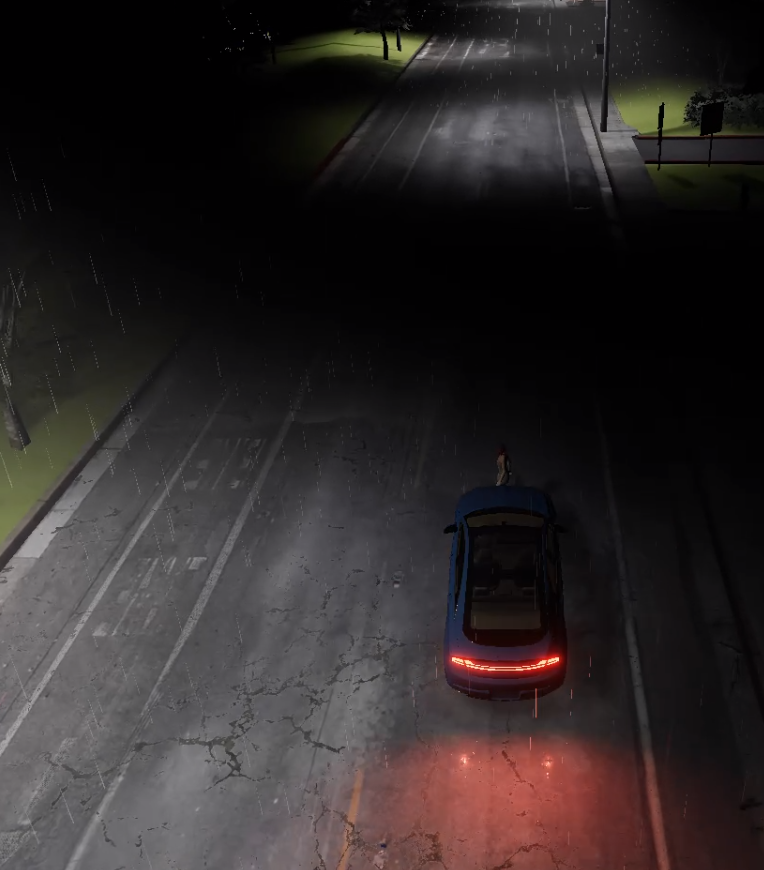}
  \caption{Collision between pedestrian and the ego vehicle on a rainy night.}
  \label{fig:collision}
\end{figure}

In \texttt{ScenarioGenerator}, the user-controllable parameters for test scenario creation and manipulation are as follows:
\begin{itemize}
	\item Initial JSON file created by SVL Visual Scenario Editor.
	\item Test case generation strategy, which is used for scenario generation. Currently, Differential Evolution, Powell Optimization, Genetic Algorithm, and random generation strategies are supported. Meanwhile, the capability of replaying a scenario is also supported by passing the JSON file and setting the action to \textit{replay}.
	A specific noise vector in combination with replay action can also be used. In this mode, in addition to all the previous parameters, a specific noise vector is given to be played. 
	\item The ego vehicle destination. 
	\item Acceptable range of changes in the values for the \textit{position of each waypoint} $(x,z)$.
	\item Acceptable range of changes in the color of each agent (r, g, b).
	\item Acceptable range of changes in the weather in the simulation (e.g., rain, fog, wetness, cloudiness, road damages).
	\item Acceptable range of changes in the time of day.
	\item Acceptable range of changes in the speed of each agent.
\end{itemize}

\gregDone{In a test case, the generated noise vector is used to impose changes to the position of each waypoint, the color of each agent, the weather, the time of day, and the speed of each agent. The base scenario defines a value for each of these parameters. The user-controllable parameters are used to constrain the range of changes made by the noise vector between minimum and maximum values, as discussed in Section~\ref{sec:TCGenerationSBST}.}

}



\report{
\section{Results and Discussion}\label{sec:results}

\hamidDone{\textit{Detected Failures.} Table \ref{table:result} shows the result of 208 executions of the simulator as described in the previous section. Fig. \ref{fig:collision} also shows a sample of the generated test scenarios leading to a collision between the pedestrian and the ego vehicle.
The number of traffic collisions and the total distance that the ego-vehicle traveled increased when the genetic algorithm is used. However, the total distance of objects from ego vehicle during scenario execution decreased. This can be explained by considering the fact that the vehicle has traveled a longer total distance.}}

\report{
\begin{figure}
  \centering
  \includegraphics[width=.95\columnwidth]{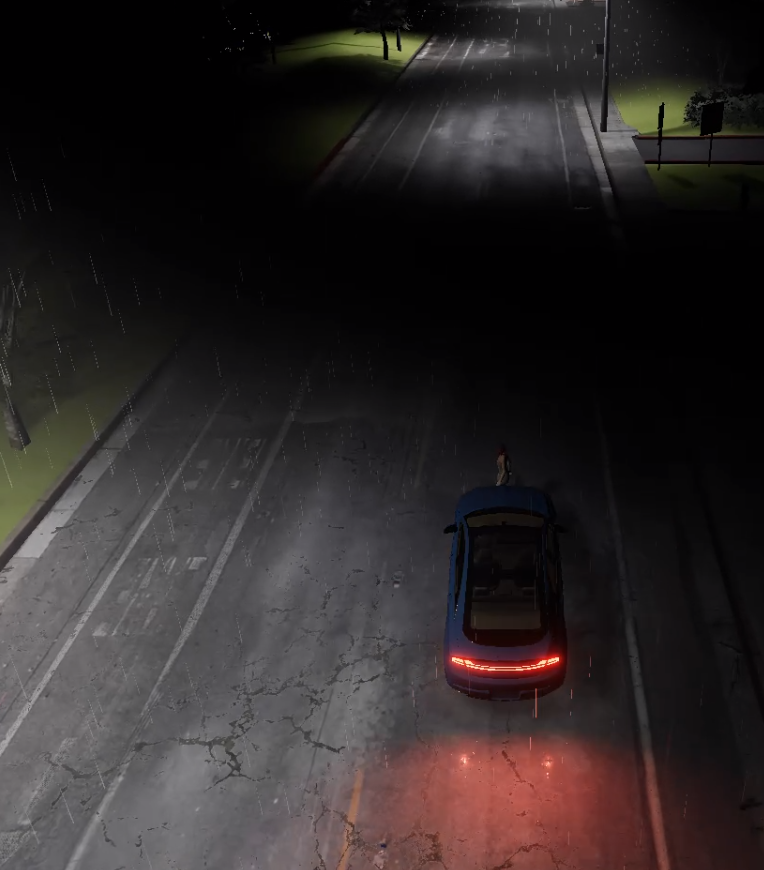}
  \caption{Collision between pedestrian and the ego vehicle in a rainy night.}
  \label{fig:collision}
\end{figure}}

\report{
The raw data (JSON scenarios, the script report and SVL simulatior analysis and videos) for both experiments can be found in \href{https://drive.google.com/file/d/1JJN90mgTfv-dIXvdV38mJRvsr4Nu2OOM/view?usp=sharing}{this link}.}

\report{
\begin{table}
\centering
\caption{Number of collisions and objective values}
\begin{tabular}{ |p{3.5cm}||p{1.5cm}|p{2.2cm}|  }
 \hline
 Objectives & Random  & Genetic algorithm \\
 \hline
 Number of collisions   & 8     & 16 \\
 Total journey distance &   6052.53m  & 6104.97 \\
 Total distance from ego & 67170.16m & 72949.39 \\
 \hline
\end{tabular}
\label{table:result}
\end{table}
}

\report{
\textit{Failure Diversity:} As we discussed earlier, we use pairwise Euclidean distance between the the noise vectors, to show diversity between the generated test cases leading to failures. Table \ref{table:Failure diversity_report} shows the range of average pairwise Euclidean distance for the failure test cases in GA and random testing. In this regard, the proposed GA-based technique promotes more diversity between the generated failure test cases than random testing.} 

\report{
\begin{table}
\centering
\caption{Failure diversity in GA and random testing}
\begin{tabular}{ |m{3cm}|m{1.9cm}|m{1.9cm}|}
\hline
\textbf{}& \textbf{GA-based test case generation}& \textbf{Random testing}\\
\hline
\hline
\textbf{\small Range of average pairwise Euclidean distance for test cases} & $4.1-4.7$ & $3.2-4.2$\\[5pt] 
\hline
\end{tabular}
\label{table:Failure diversity_report}
\end{table}
}
\report{
\subsection{Discovered problems and statistics}\label{Sec:problems}
Reproducible scenarios are not possible. The simulation execution does not produce identical results with identical input parameters

During the experimentation we noticed that many of failures that are captured are not completely reproducible. It implies that the possibility of reproducible test scenarios is excluded, since simulation execution does not produce identical results upon identical input parameters and same configuration setup.
We believe it is partially because Apollo is not deterministic and while it is possible to use simulator's clock sensor to synchronize time between SVL Simulator and Apollo, this not enough to achieve reproducible simulations.}



\paper{

\section{Results and Discussion}\label{sec:results}

This section presents the experimental results and assesses the proposed test case generation compared to the random testing with regard to the quality criteria.

\textit{Detected Failures:} Fig. \ref{fig:failure_data}(a) shows the number of detected failures (test cases leading to collisions) by the GA-based test case generation and random testing. The proposed GA-based technique trigger twice as many failures as random testing on the same configuration and test budget, and consequently, in this regard, works more effectively. Fig. \ref{fig:collision} also shows a sample of a generated test scenario leading to a collision between the pedestrian and the ego vehicle. 

In order to investigate the characteristics of the detected failures, we can examine the values of two of the objectives in the objective function---$ego\_agents\_distance$ and $journey\_distance$. These can show the characteristics of the detected failures. Fig. \ref{fig:failure_data}(b) and (c) show the average values of the two objectives in failure-revealing test cases for both techniques. These average values do not differ significantly between the two approaches. This indicates that the GA reveals more failures, but the failures revealed by the two techniques fall in similar objective ranges. However, both distances are somewhat higher in the GA---i.e., the GA generates tests with slightly longer journey distances and a slightly higher distance from the ego car. These tests may be somewhat more interesting for revealing errors in the ego car functionality, as---for example---a longer distance between the ego car and a pedestrian should offer more time to make corrections. In future work, we will examine failing scenarios more closely and discuss them with domain experts.

\textit{Failure Diversity:} We use pairwise Euclidean distance between the noise vectors to show diversity between the failure-revealing test cases. Fig. \ref{fig:GADiversity} and \ref{fig:RandomDiversity} show the average pairwise Euclidean distance for each of the failure test cases generated by GA and random testing respectively. The average pairwise Euclidean distance refers to the average difference between a test case and the other test cases. Table \ref{table:Failure diversity} shows the range of average pairwise Euclidean distance for the failure-revealing test cases from the GA and random testing. In this regard, the GA technique also promotes more diversity between generated failure-revealing test cases than random testing.    

\begin{figure}[!t]
  \centering
  \includegraphics[width=.9\columnwidth, height = 5cm]{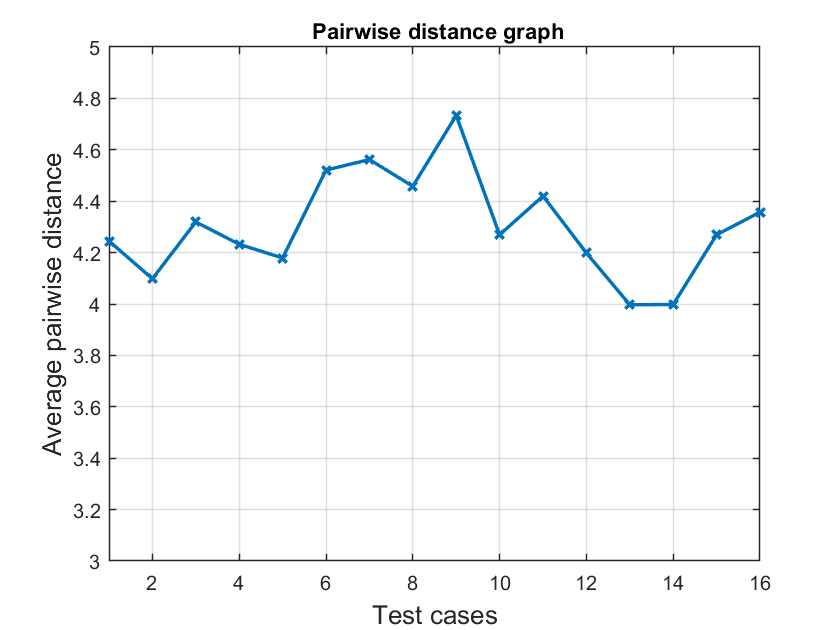}
  \caption{Diversity of failure-revealing test cases generated by the GA.} 
  \label{fig:GADiversity}
\end{figure}

\begin{figure}[!t]
  \centering
  \includegraphics[width=.9\columnwidth, height = 5cm]{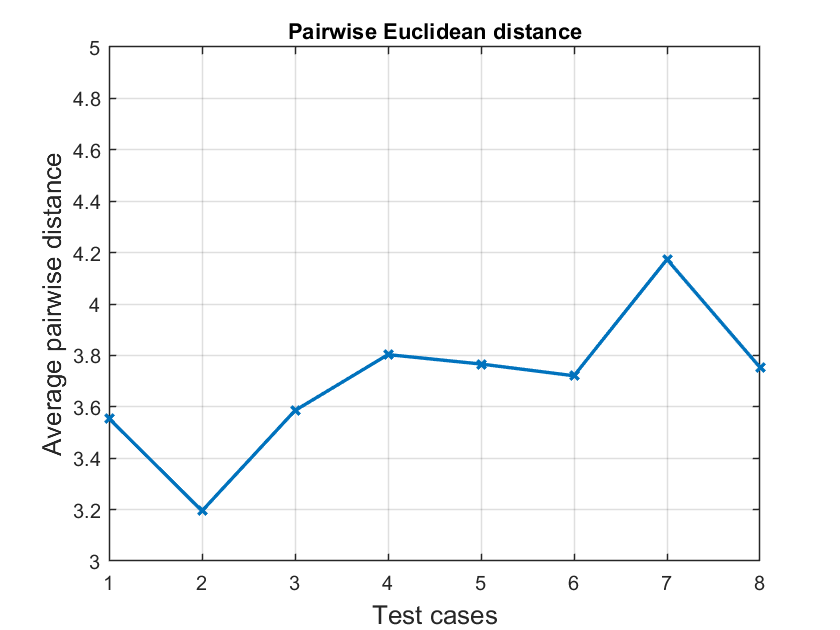}
  \caption{Diversity of failure-revealing test cases generated by random testing.} 
  \label{fig:RandomDiversity}
\end{figure}

\begin{table}[!t]
\centering
\caption{Failure diversity, shown as the range in the average pairwise Euclidean distance for test cases.}
\begin{tabular}{|l|r|r|}
\cline{2-3}
\multicolumn{1}{c|}{}& \textbf{Genetic Algorithm}& \textbf{Random}\\
\hline
\textbf{Range of Euclidean Distances} & $4.1-4.7$ & $3.2-4.2$\\
\hline
\end{tabular}
\label{table:Failure diversity}
\end{table}
}

\paper{
\subsection{Threats to Validity}\label{sec:Threats}
Some of the main sources of threats to validity of the experimental results are as follows:

\smallskip\noindent\textbf{Internal Validity:} 
During the experiment, we noticed that many of the failures that are captured are not completely reproducible. 
In fact, the simulation execution often does not produce identical results given identical input parameters and configuration setup.
One of the main reasons is that Apollo does not function in a deterministic manner. We tried to mitigate the effects of this by reporting average values from the experiments, and conducting the experiments in a controlled manner, i.e., using the same experimental setup and keeping the user-controllable parameters fixed between executions. Another source of threat is the fact that as the simulator runs a large number of test cases, the simulations become slower and less responsive probably due to performance bottlenecks. 

\smallskip\noindent\textbf{External Validity:} We have focused on a single scenario. As we have used a generic data structure consisting of variables scaled in a certain range, i.e., the noise vector with variables within the range $[-1, 1]$, we believe that the representation model and test case generation approach could be used for simulation-based testing of more complex scenes and other use cases. However, the variables in the noise vector might need to be modified (e.g., extended) for different use cases.   
}

\paper{
\section{Related Work} \label{sec:relatedwork}
Simulators as a form of digital twins play a key role for different purposes in testing and verification, control and monitoring, and improvement of cyber-physical systems (CPS). For ADAS and autonomous-driving cars, this is even more significant and there is a higher demand for high-fidelity simulators. Simulation-based testing is one of the most effective approaches for system-level testing of ADAS and acts as a suitable complementary solution to on-road testing, since it provides the possibility for early stage testing, capturing critical corner test scenarios and enabling inexpensive testing. Field testing of such systems is expensive, inefficient and even dangerous, in some cases. Recently, various simulators such as those ones using physics-based models, e.g., SVL simulator \cite{SVLSimulator}, PreScan \cite{PreScan} and Pro-SiVIC \cite{ProSivic_belbachir2012simulation} or the ones relying on game engines, e.g., BeamNG \cite{BeamNG} and CARLA \cite{dosovitskiy2017carla}, have been developed to meet the need for realistic simulation of the functions in autonomous driving. 

Accordingly, various system-level testing approaches relying on the simulators have been proposed in the recent years. One of the common intended purposes in those studies is generating critical test cases (scenarios) that lead the system to fail. This is a challenging problem, due to the large search space of input parameters in these systems. Covering all possible simulation test scenarios is not feasible in practice. Therefore, in this regard SBST techniques have been widely used to generate effective test simulation scenarios for those systems. In recent studies, multi-objective search algorithms like NSGA-II \cite{ben_abdessalem_testing_2016}, many-objective algorithms like MOSA \cite{panichella2015reformulating} using a combination of different objectives based on branch coverage and failure-based heuristics \cite{abdessalem2018testing}, and learnable evolutionary algorithms \cite{abdessalem2018testinglearnable} have been used to generate critical test cases leading to violations of safety requirements in autonomous driving cars. Moreover, there have also been studies focusing on the role of simulators and the type of test data. In \cite{haq2020comparing} a comparison between testing of DNN-based ADAS using real-world and simulator-generated data is conducted and it is also showed that how on-line and off-line testing of these systems can differ and meanwhile complement each other. Markus et al. studied the consistency between the results obtained from two different simulators and investigated whether the obtained results could be mutually reproducible in both simulators \cite{borg2021digital}.
}


\section{Conclusion and Future Work} \label{sec:summary}
Efficient and effective test case generation for use in virtual environments is essential for testing AI-based automotive systems. In this \paper{paper}\report{report}, we presented a SBST approach to generate test scenarios that lead to detection of failures and safety violations of the Baidu Apollo pedestrian emergency braking system. We have made three primary observations. First, our results show that the proposed GA-based test case generation is more effective than random testing, i.e., it is more effective in generating failure revealing test cases and provides higher diversity between the generated test cases compared to random testing.
Second, unfortunately, many of the captured failures could not be reproduced given the same configuration and user-controlled parameters due to the non-deterministic nature of Apollo.
Third, we see great potential in simulation-based testing of different functions of autonomous driving systems using SVL simulator and Baidu Apollo. In future work, we will broaden the scope of the research into additional safety scenarios. We will also extend SBST approaches with machine learning-based techniques (e.g., reinforcement learning) for test case generation in system-level testing of ADAS. We are also interested in the use of Generative Adversarial Networks (GANs) as a technique for enabling the discovery of failure-revealing test cases.

\section*{Acknowledgment}
This project has received funding from the ECSEL Joint Undertaking (JU) under grant agreement No 876852 (VALU3S). Furthermore, this work received support from the ITEA3 European IVVES project (\url{https://itea3.org/project/ivves.html}) and the SMILE~III project financed by Vinnova, FFI, Fordonsstrategisk forskning och innovation under the grant numbers 2019-05871 and the AIQ Meta-Testbed project funded by Kompetensfonden at Campus Helsingborg, Lund University, Sweden. Additional support was provided under Vetenskapsr{\aa}det grant 2019-05275.
\hamidDone{The authors would like to thank INFOTIV AB for their support and cooperation.}

\bibliographystyle{IEEEtran}
\bibliography{competition_refs}

\begin{thebibliography}{10}
\providecommand{\url}[1]{#1}
\csname url@samestyle\endcsname
\providecommand{\newblock}{\relax}
\providecommand{\bibinfo}[2]{#2}
\providecommand{\BIBentrySTDinterwordspacing}{\spaceskip=0pt\relax}
\providecommand{\BIBentryALTinterwordstretchfactor}{4}
\providecommand{\BIBentryALTinterwordspacing}{\spaceskip=\fontdimen2\font plus
\BIBentryALTinterwordstretchfactor\fontdimen3\font minus
  \fontdimen4\font\relax}
\providecommand{\BIBforeignlanguage}[2]{{%
\expandafter\ifx\csname l@#1\endcsname\relax
\typeout{** WARNING: IEEEtran.bst: No hyphenation pattern has been}%
\typeout{** loaded for the language `#1'. Using the pattern for}%
\typeout{** the default language instead.}%
\else
\language=\csname l@#1\endcsname
\fi
#2}}
\providecommand{\BIBdecl}{\relax}
\BIBdecl

\bibitem{ec_ai_def}
``A definition of artificial intelligence: Main capabilities and scientific
  disciplines,'' High-Level Expert Group on Artificial Intelligence, Brussels,
  Belgium, Tech. Rep., 2018.

\bibitem{borg2019safely}
M.~Borg, C.~Englund, K.~Wnuk, B.~Duran, C.~Lewandowski, S.~Gao, Y.~Tan,
  H.~Kaijser, H.~L{\"o}nn, and J.~T{\"o}rnqvist, ``Safely entering the deep: A
  review of verification and validation for machine learning and a challenge
  elicitation in the automotive industry,'' \emph{Journal of Automotive
  Software Engineering}, vol.~1, no.~1, pp. 1--19, 2019.

\bibitem{ec_ai_strategy}
``Communucation from the commission to the european parliament, the european
  council, the europan economic and social commitee and the commitee of the
  regions - artificial intelligence for europe,'' Eurpean Commision, Brussels,
  Belgium, Tech. Rep., 2018.

\bibitem{zhang2020machine}
J.~M. Zhang, M.~Harman, L.~Ma, and Y.~Liu, ``Machine learning testing: Survey,
  landscapes and horizons,'' \emph{IEEE Transactions on Software Engineering},
  2020.

\bibitem{riccio2020testing}
V.~Riccio, G.~Jahangirova, A.~Stocco, N.~Humbatova, M.~Weiss, and P.~Tonella,
  ``Testing machine learning based systems: a systematic mapping,''
  \emph{Empirical Software Engineering}, vol.~25, no.~6, pp. 5193--5254, 2020.

\bibitem{bock2019status}
F.~Bock, C.~Sippl, S.~Siegl, and R.~German, ``Status report on automotive
  software development,'' in \emph{Automotive Systems and Software
  Engineering}.\hskip 1em plus 0.5em minus 0.4em\relax Springer, 2019, pp.
  29--57.

\bibitem{koopman2016challenges}
P.~Koopman and M.~Wagner, ``Challenges in autonomous vehicle testing and
  validation,'' \emph{SAE International Journal of Transportation Safety},
  vol.~4, no.~1, pp. 15--24, 2016.

\bibitem{international_organization_for_standardization_road_2019}
``Road {Vehicles} - {Safety} of the {Intended} {Functionality},'' International
  Organization for Standardization, Tech. Rep. ISO/PAS 21448:2019, 2019.

\bibitem{mcminn2011search}
P.~McMinn, ``Search-based software testing: Past, present and future,'' in
  \emph{2011 IEEE Fourth International Conference on Software Testing,
  Verification and Validation Workshops}, 2011, pp. 153--163.

\bibitem{ben_abdessalem_testing_2016}
R.~Ben~Abdessalem, S.~Nejati, L.~C. Briand, and T.~Stifter, ``Testing advanced
  driver assistance systems using multi-objective search and neural networks,''
  in \emph{Proc. of the 31st {IEEE}/{ACM} {International} {Conference} on
  {Automated} {Software} {Engineering}}, 2016, pp. 63--74.

\bibitem{ben_abdessalem_testing_2018}
------, ``Testing vision-based control systems using learnable evolutionary
  algorithms,'' in \emph{Proc. of the 40th {International} {Conference} on
  {Software} {Engineering}}, 2018, pp. 1016--1026.

\bibitem{gambi_generating_2019}
A.~Gambi, T.~Huynh, and G.~Fraser, ``Generating effective test cases for
  self-driving cars from police reports,'' in \emph{Proc. of the 27th {ACM}
  {Joint} {Meeting} on {European} {Software} {Engineering} {Conference} and
  {Symposium} on the {Foundations} of {Software} {Engineering}}, 2019, pp.
  257--267.

\bibitem{borg2021digital}
M.~Borg, R.~B. Abdessalem, S.~Nejati, F.-X. Jegeden, and D.~Shin, ``Digital
  twins are not monozygotic--cross-replicating adas testing in two
  industry-grade automotive simulators,'' in \emph{2021 14th IEEE Conference on
  Software Testing, Verification and Validation (ICST)}.\hskip 1em plus 0.5em
  minus 0.4em\relax IEEE, 2021, pp. 383--393.

\bibitem{Moghadam2021Deeper}
M.~H. Moghadam, M.~Borg, and S.~J. Mousavirad, ``Deeper at the sbst 2021 tool
  competition: {ADAS} testing using multi-objective search,'' in \emph{2021
  14th Intl. Workshop on Search-Based Software Testing (SBST)}.\hskip 1em plus
  0.5em minus 0.4em\relax IEEE, 2021.

\bibitem{deb2002fast}
K.~Deb, A.~Pratap, S.~Agarwal, and T.~Meyarivan, ``A fast and elitist
  multiobjective genetic algorithm: {NSGA-II},'' \emph{IEEE transactions on
  evolutionary computation}, vol.~6, no.~2, pp. 182--197, 2002.

\bibitem{SVLSimulator}
{LG Electronics}, ``{SVL Simulator},'' {\url{https://www.svlsimulator.com/}},
  Retrieved July, 2021.

\bibitem{PreScan}
{TASS International}, ``{. PreScan Simulator},''
  {\url{https://tass.plm.automation.siemens.com/prescan-overview}}, Retrieved
  July, 2021.

\bibitem{ProSivic_belbachir2012simulation}
A.~Belbachir, J.-C. Smal, J.-M. Blosseville, and D.~Gruyer, ``Simulation-driven
  validation of advanced driving-assistance systems,'' \emph{Procedia-Social
  and Behavioral Sciences}, vol.~48, pp. 1205--1214, 2012.

\bibitem{BeamNG}
{BeamNG GmbH.}, ``{BeamNG.research},'' {\url{https://beamng.gmbh/research/}},
  Retrieved July, 2021.

\bibitem{dosovitskiy2017carla}
A.~Dosovitskiy, G.~Ros, F.~Codevilla, A.~Lopez, and V.~Koltun, ``Carla: An open
  urban driving simulator,'' in \emph{Conference on robot learning}.\hskip 1em
  plus 0.5em minus 0.4em\relax PMLR, 2017, pp. 1--16.

\bibitem{panichella2015reformulating}
A.~Panichella, F.~M. Kifetew, and P.~Tonella, ``Reformulating branch coverage
  as a many-objective optimization problem,'' in \emph{2015 IEEE 8th
  international conference on software testing, verification and validation
  (ICST)}.\hskip 1em plus 0.5em minus 0.4em\relax IEEE, 2015, pp. 1--10.

\bibitem{abdessalem2018testing}
R.~B. Abdessalem, A.~Panichella, S.~Nejati, L.~C. Briand, and T.~Stifter,
  ``Testing autonomous cars for feature interaction failures using
  many-objective search,'' in \emph{2018 33rd IEEE/ACM International Conference
  on Automated Software Engineering (ASE)}.\hskip 1em plus 0.5em minus
  0.4em\relax IEEE, 2018, pp. 143--154.

\bibitem{abdessalem2018testinglearnable}
R.~B. Abdessalem, S.~Nejati, L.~C. Briand, and T.~Stifter, ``Testing
  vision-based control systems using learnable evolutionary algorithms,'' in
  \emph{2018 IEEE/ACM 40th International Conference on Software Engineering
  (ICSE)}.\hskip 1em plus 0.5em minus 0.4em\relax IEEE, 2018, pp. 1016--1026.

\bibitem{haq2020comparing}
F.~U. Haq, D.~Shin, S.~Nejati, and L.~C. Briand, ``Comparing offline and online
  testing of deep neural networks: An autonomous car case study,'' in
  \emph{2020 IEEE 13th International Conference on Software Testing, Validation
  and Verification (ICST)}.\hskip 1em plus 0.5em minus 0.4em\relax IEEE, 2020,
  pp. 85--95.

\end{thebibliography}

\end{document}